\documentclass[10pt,twocolumn,letterpaper]{article}

\usepackage{iccv}
\usepackage{times}
\usepackage{epsfig}
\usepackage{graphicx}
\usepackage{amsmath}
\usepackage{amssymb}

\usepackage{multirow}
\usepackage{booktabs} % To thicken table lines

\usepackage[pagebackref=true,breaklinks=true,letterpaper=true,colorlinks,bookmarks=false]{hyperref}

\iccvfinalcopy % *** Uncomment this line for the final submission

\def\iccvPaperID{654} % *** Enter the ICCV Paper ID here

\ificcvfinal\fi
\begin{document}
	
	%%%%%%%%% TITLE
	\title{LightTrack: A Generic Framework for Online Top-Down Human Pose Tracking}
	
	\author{Guanghan Ning\textsuperscript{1}, Heng Huang\textsuperscript{1, 2}\\
		%\textsuperscript{1}JD Finance America, Mountain View, CA, USA\\
		\textsuperscript{1}JD Digits, Mountain View, CA, USA\\
		\textsuperscript{2} University of Pittsburgh, Pittsburgh, PA, USA\\
		%\{guanghan.ning, heng.huang\}@jd.com
	}
	
	\maketitle
	%\thispagestyle{empty}

	%%%%%%%%% ABSTRACT
	\begin{abstract}
		In this paper, we propose a novel effective light-weight framework, called as LightTrack, for online human pose tracking. The proposed framework is designed to be generic for top-down pose tracking and is faster than existing online and offline methods. 
		Single-person Pose Tracking (SPT) and Visual Object Tracking (VOT) are incorporated into one unified functioning entity, easily implemented by a replaceable single-person pose estimation module. Our framework unifies single-person pose tracking with multi-person identity association and sheds first light upon bridging keypoint tracking with object tracking.
		We also propose a Siamese Graph Convolution Network (SGCN) for human pose matching as a Re-ID module in our pose tracking system. In contrary to other Re-ID modules, we use a graphical representation of human joints for matching. The skeleton-based representation effectively captures human pose similarity and is computationally inexpensive. It is robust to sudden camera shift that introduces human drifting.
		To the best of our knowledge, this is the first paper to propose an online human pose tracking framework in a top-down fashion. The proposed framework is general enough to fit other pose estimators and candidate matching mechanisms.
		Our method outperforms other online methods while maintaining a much higher frame rate, and is very competitive with our offline state-of-the-art.
		We make the code publicly available at: \href{https://github.com/Guanghan/lighttrack}{https://github.com/Guanghan/lighttrack}.
	\end{abstract}

	%%%%%%%%% BODY TEXT
	\section{Introduction}
	
	\begin{figure*}
		\centering
		\includegraphics[width=0.9\linewidth]{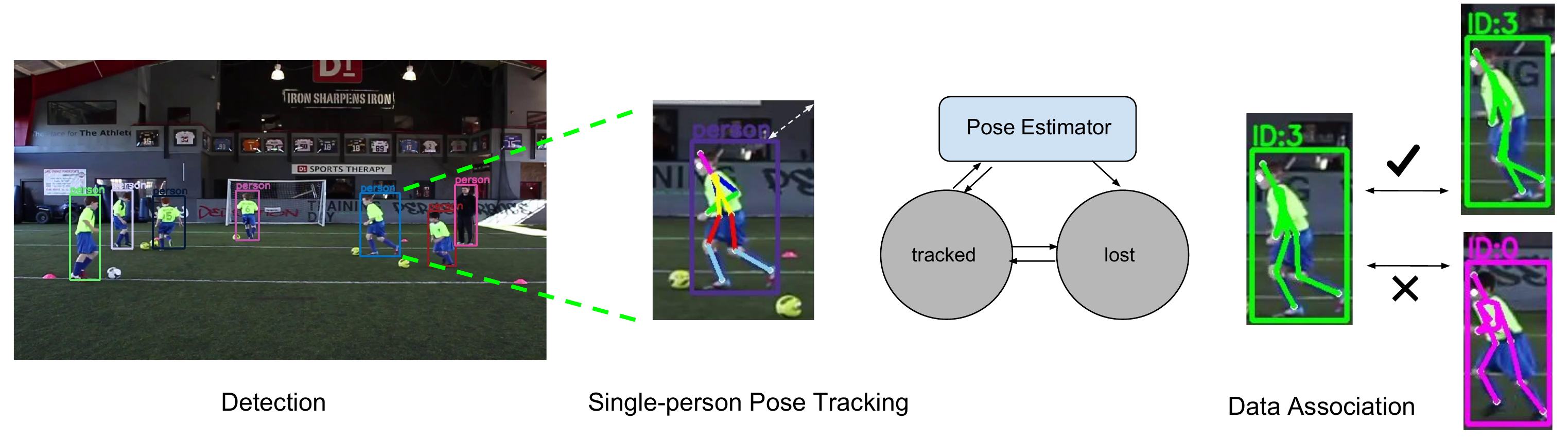}
		\caption{Overview of the proposed online pose tracking framework. We detect human candidates in the first frame, then track each candidate's position and pose by a single-person pose estimator. When a target is lost, we perform detection for this frame and data association with a graph convolution network for skeleton-based pose matching. We use skeleton-based pose matching because visually similar candidates with different identities may confuse visual classifiers. Extracting visual features can also be computationally expensive in an online tracking system. Pose matching is considered because we observe that in two adjacent frames, the location of a person may drift away due to sudden camera shift, but the human pose will stay almost the same as people usually cannot act that fast.} 
		\label{fig:overview}
		\vspace{-.2in}
	\end{figure*}
	
	Pose tracking is the task of estimating multi-person human poses in videos and assigning unique instance IDs for each keypoint across frames. Accurate estimation of human keypoint-trajectories is useful for human action recognition, human interaction understanding, motion capture and animation, \emph{etc}.
	Recently, the publicly available PoseTrack dataset \cite{iqbal2017posetrack, andriluka2018posetrack} and MPII Video Pose dataset \cite{insafutdinov2017arttrack} have pushed the research on human motion analysis one step further to its real-world scenario.
	Two PoseTrack challenges have been held. 
	However, most existing methods are offline hence lacking the potential to be real-time. More emphasis has been put on the \textit{Multi-Object Tracking Accuracy} (MOTA) criterion compared to the \textit{Frame Per Second} (FPS) criterion. 
	Existing offline methods divide the tasks of human detection, candidate pose estimation, and identity association into sequential stages.
	In the procedure, multi-person poses are estimated across frames within a video. Based on the pose estimation results, the pose tracking outputs are computed via solving an optimization problem. It requires the poses of future frames to be pre-computed, or at least for the frames within some range. %Pose tracking has been treated as an optimization problem by the offline methods. 
	%Hungarian algorithm is the most commonly used offline method.
	
	In this paper, we propose a novel effective light-weight framework for pose tracking. It is designed to be generic, top-down (\emph{i.e.}, pose estimation is performed after candidates are detected), and truly online.  
	The proposed framework unifies single-person pose tracking with multi-person identity association. It sheds first light on bridging keypoint tracking with object tracking.
	To the best of our knowledge, this is the first paper to propose an online pose tracking framework in a top-down fashion. The proposed framework is general enough to fit other pose estimators and candidate matching mechanisms. Thus, if individual component is further improved in the future, our framework will be faster and/or more accurate.

	In contrast to \textit{Visual Object Tracking} (VOT) methods, in which the visual features are implicitly represented by kernels or CNN feature maps, we track each human pose by recursively updating the bounding box and its corresponding pose in an explicit manner. The bounding box region of a target is inferred from the explicit features, \emph{i.e.}, the human keypoints. Human keypoints can be considered as a series of special visual features. 
	The advantages of using pose as explicit features include: (1) The explicit features are human-related and interpretable, and have very strong and stable relationship with the bounding box position. Human pose enforces direct constraint on the bounding box region. (2) The task of pose estimation and tracking requires human keypoints be predicted in the first place. Taking advantage of the predicted keypoints is efficient in tracking the ROI region, which is almost free. This mechanism makes the online tracking possible. (3) It naturally keeps the identity of the candidates, which greatly alleviates the burden of data association in the system. Even when data association is necessary, we can re-use the pose features for skeleton-based pose matching.
	\textit{Single Pose Tracking} (SPT) and \textit{Single Visual Object Tracking} (VOT) are thus incorporated into one unified functioning entity, easily implemented by a replaceable single-person human pose estimation module.
	
	Our contributions are in three-fold: (1) We propose a general online pose tracking framework that is suitable for top-down approaches of human pose estimation. Both human pose estimator and Re-ID module are replaceable. In contrast to \textit{Multi-Object Tracking} (MOT) frameworks, our framework is specially designed for the task of pose tracking. To the best of our knowledge, this is the first paper to propose an online human pose tracking system in a top-down fashion. 
	(2) We propose a \textit{Siamese Graph Convolution Network} (SGCN) for human pose matching as a Re-ID module in our pose tracking system. Different to existing Re-ID modules, we use a graphical representation of human joints for matching. The skeleton-based representation effectively captures human pose similarity and is computationally inexpensive. It is robust to sudden camera shift that introduces human drifting.
	(3) We conduct extensive experiments with various settings and ablation studies. Our proposed online pose tracking approach outperforms existing online methods and is competitive to the offline state-of-the-arts but with much higher frame rates. We make the code publicly available to facilitate future research.

	%------------------------------------------------------------------------
	\section{Related Work}
	\subsection{Human Pose Estimation and Tracking} 
	
	\textit{Human Pose Estimation} (HPE) has seen rapid progress with the emergence of CNN-based methods \cite{wei2016convolutional, newell2016stacked, yang2017learning, ke2018multi}. 
	%This task is relatively easier, because human candidates are cropped and centered in the image patch. 
	The most widely used datasets, \emph{e.g.}, MPII \cite{andriluka14cvpr} and LSP \cite{Johnson10}, are saturated with methods that achieve 90\% and higher accuracy.
	Multi-person human pose estimation is more realistic and challenging, and has received increasing attentions with the hosting of COCO keypoints challenges \cite{lin2014microsoft} since 2017. Existing methods can be classified into top-down and bottom-up approaches. The top-down approaches \cite{fang2017rmpe,papandreou2017towards,he2017mask} rely on the detection module to obtain human candidates and then applying single-person pose estimation to locate human keypoints. The bottom-up methods  \cite{cao2016realtime, xia2017joint, newell2016associative}
	%, on the other hand, 
	detect human keypoints from all potential candidates and then assemble these keypoints into human limbs for each individual based on various data association techniques. 
	The advantage of bottom-up approaches is their excellent trade-off between estimation accuracy and computational cost because the cost is nearly invariant to the number of human candidates in the image. 
	In contrast, the advantage of top-down approaches is their capability in disassembling the task into multiple comparatively easier tasks, \emph{i.e.}, object detection and single-person pose estimation. The object detector is expert in detecting hard (usually small) candidates, so that the pose estimator will perform better with a focused regression space.
	Pose tracking is a new topic that is primarily introduced by the PoseTrack dataset \cite{iqbal2017posetrack, andriluka2018posetrack} and MPII Video Pose dataset \cite{insafutdinov2017arttrack}. 
	%It is one step closer to the real-world problem for human motion analysis. 
	The task is to estimate human keypoints and assign unique IDs to each keypoint at instance-level across frames in videos. 
	%Accurate trajectory estimation of human keypoints is useful in human action recognition and human interaction understanding.
	A typical top-down but offline method was introduced in \cite{insafutdinov2017arttrack}, where pose tracking is transformed into a minimum cost multi-cut problem with a graph partitioning formulation.

	\subsection{Object Detection vs. Human Pose Estimation} 
	Earlier works in object detection regress visual features into bounding box coordinates. HPE, on the other hand, usually regresses visual features into heatmaps, each channel representing a human joint. 
	Recently, research in HPE has inspired many works on object detection \cite{zhou2019bottomup, law2018cornernet, Man+18}. These works predict heatmaps for a set of special keypoints to infer detection results (bounding boxes).
	Based on this motivation,
	we propose to predict human keypoints to infer bounding box regions. Human keypoints are a special set of keypoints to represent detection of the human class only.
	
	\subsection{Multi-Object Tracking}
	MOT aims to estimate trajectories of multiple objects by finding target locations while maintaining their identities across frames.
	Offline methods use both past and future frames to generate trajectories while online methods only exploit information that is available until the current frame.
	An online MOT pipeline \cite{zhu2018online} was presented with applying a single object tracker to keep tracking each target given these target detections in each frame. The target state is set as tracked until the tracking result becomes unreliable. The target is then regarded as lost, and data association is performed to compute the similarity between the track-let and detections. 
	Our proposed online pose tracking framework also tracks each target (with corresponding keypoints) individually while keeping their identities, and performs data association when target is lost. However, our framework is distinct in several aspects: (a) the detection is generated by object detector only at key frames, therefore not necessarily provided at each frame. It can be provided scarcely; (b) the single object tracker is actually a pose estimator that predicts keypoints based on an enlarged region.
	
	\subsection{Graphical Representation for Human Pose} 
	It is recently studied in \cite{stgcn2018aaai} on how to effectively model dynamic skeletons with a specially tailored graph convolution operation. The graph convolution operation turns human skeletons into spatio-temporal representation of human actions. Inspired by this work, we propose to employ GCN to encode spatial relationship among human joints into a latent representation of human pose. %, which conceptually “summarizes” the pose. 
	The representation aims to robustly encode the pose, which is invariant to human location or view angle. 
	%If the same 3D pose is mapped into different planes, the corresponding 2D poses should result in similar encodings in the latent space. 
	We measure similarities of such encodings for the matching of human poses. 
	
	%-------------------------------------------------------------------------
	\section{Proposed Method}
	
	\subsection{Top-Down Pose Tracking Framework}
	%In this paper, we look into online human pose tracking in a top-down fashion.
	We propose a novel top-down pose tracking framework.
	It has been proved that human pose can be employed for better inference of human locations \cite{liu2018posehd}. We observe that, in a top-down approach, accurate human locations also ease the estimation of human poses. We further study the relationships between these two levels of information:
	(1)	Coarse person location can be distilled into body keypoints by a single-person pose estimator.
	(2)	The position of human joints can be straightforwardly used to indicate rough locations of human candidates. 
	(3)	Thus, recurrently estimating one from the other is  a feasible strategy for \textit{Single-person Pose Tracking} (SPT).  
	
	However, it is not a good idea to merely consider the \textit{Multi-target Pose Tracking} (MPT) problem as a repeated SPT problem for multiple individuals. Because certain constraints need to be met, \emph{e.g.}, in a certain frame, two different IDs should not belong to the same person; neither  two candidates should share the same identity. A better way is to track multiple individuals simultaneously and preserve/update their identities with an additional Re-ID module. 
	The Re-ID module is essential because it is usually hard to maintain correct identities all the way. It is unlikely to track the individual poses effectively across frames of the entire video. For instance, under the following scenarios, identities have to be updated: (1) some people disappear from the camera view or get occluded; (2) new candidates come in or previous candidates re-appear; (3) people walk across each other (two identities may merge into one if not treated carefully); (4) tracking fails due to fast camera shifting or zooming.
	
	In our method, we first treat each human candidate separately such that their corresponding identity is kept across the frames. In this way, we circumvent the time-consuming offline optimization procedure. In case the tracked candidate is lost due to occlusion or camera shift, we then call the detection module to revive candidates and associate them to the tracked targets from the previous frame via pose matching. In this way, we accomplish multi-target pose tracking with an SPT module and a pose matching module. 
	
	Specifically, the bounding box of the person in the upcoming frame is inferred from the joints estimated by the pose module from the current frame. We find the minimum and maximum coordinates and enlarge this ROI region by 20\% on each side. The enlarged bounding box is treated as the localized region for this person in the next frame. 
	If the average confidence score $\bar{s}$ from the estimated joints is lower than the standard $\tau_s$, it reflects that the target is lost since the joints are not likely to appear in the bounding box region. 
	The state of the target is defined as:
	\begin{equation}\label{eq:state}
	\text{state}=
	\begin{cases}
	\text{tracked}, & \text{if }  \bar{s}>\tau_s,\\
	\text{lost}, & \text{otherwise}.
	\end{cases}
	\end{equation}
	If the target is lost, we have two modes: (1) \textbf{Fixed Keyframe Interval (FKI)} mode. Neglect this target until the scheduled next key-frame, where the detection module re-generate the candidates and then associate their IDs to the tracking history. (2) \textbf{Adaptive Keyframe Interval (AKI)} mode. Immediately revive the missing target by candidate detection and identity association. 
	The advantage of FKI mode is that the frame rate of pose tracking is stable due to the fixed interval of keyframes. The advantage of AKI mode is that the average frame rate can be higher for non-complex videos.
	In our experiments, we incorporate them by taking keyframes with fixed intervals while also calling detection module once a target is lost before the arrival of the next arranged keyframe.
	The tracking accuracy is higher because when a target is lost, it is handled immediately. %The revival is no later, if not sooner, compared to FKI mode.
	
	For identity association, we propose to consider two complementary information: spatial consistency and pose consistency. We first rely on spatial consistency, \emph{i.e.}, if two bounding boxes from the current and the previous frames are adjacent, or their \textit{Intersection Over Union} (IOU) is above a certain threshold, we consider them to belong to the same target. 
	Specifically, we set the matching flag $m(t_{k}, d_{k})$ to $1$ if the maximum IOU overlap ratio $o(t_{k}, \mathcal{D}_{i, k})$ between the tracked target $t_{k} \in \mathcal{T}_{k}$ and the corresponding detection $d_{k} \in \mathcal{D}_{k}$ for key-frame $k$ is higher than the threshold $\tau_o$. Otherwise, $m(t_{k}, d_{k})$ is set as $0$:
	\begin{equation}\label{eq:matching}
	m(t_{k}, d_{k})=
	\begin{cases}
	1, & \text{if }  o(t_{k}, \mathcal{D}_{i, k}) >\tau_o,\\
	0, & \text{otherwise}.
	\end{cases}
	\end{equation}
	
	\begin{figure}
		\centering
		\includegraphics[width=0.85\linewidth]{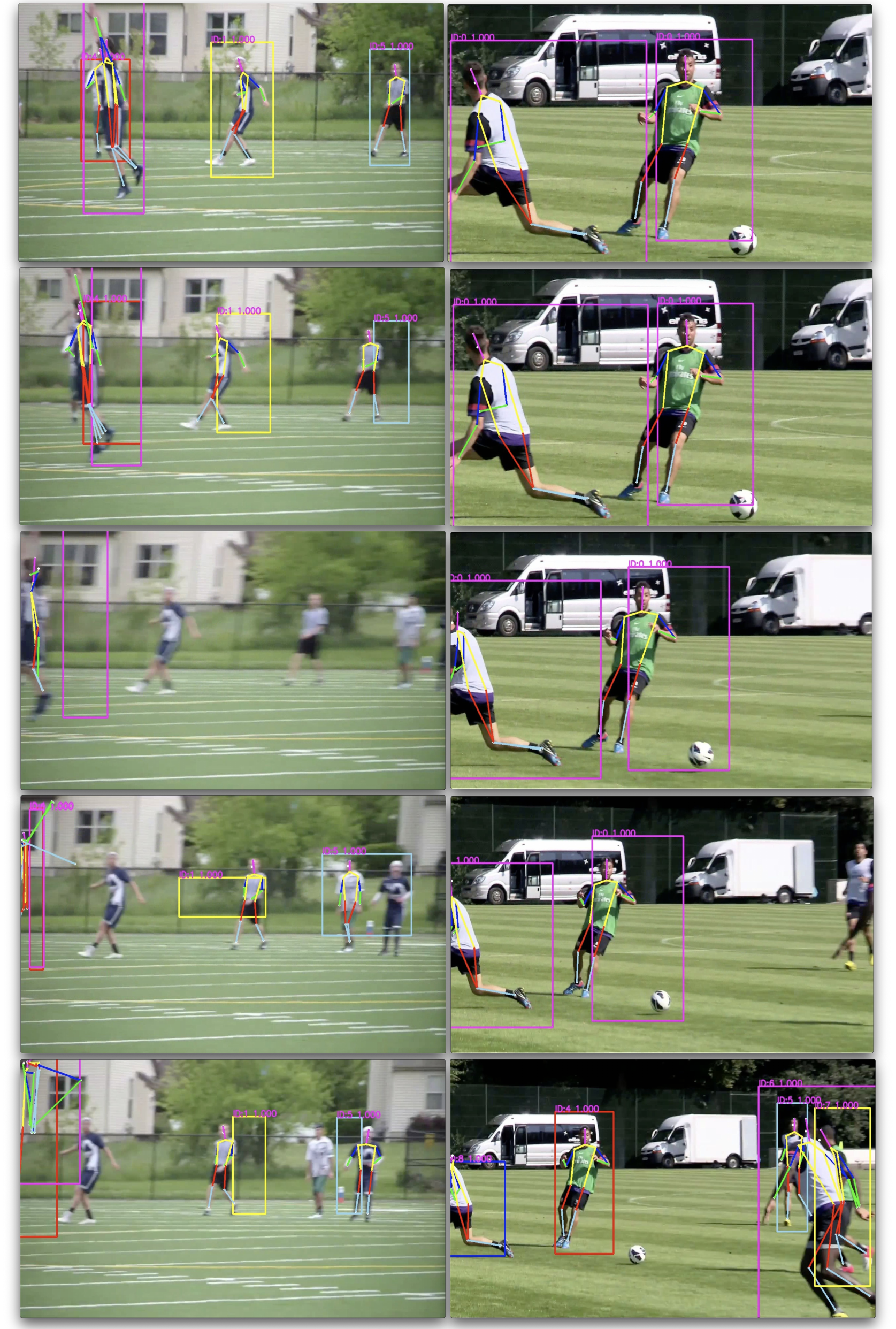}
		\caption{Sequentially adjacent frames with sudden camera shift (left frames), and sudden zooming (right frames). Each bounding box in the current frame indicates the corresponding region inferred from the human keypoints from the previous frame. The human pose in the current frame is estimated by the pose estimator. The ROI for the pose estimator is the expanded bounding box. } 
		\label{fig:camera_shift}
		\vspace{-.2in}
	\end{figure}
	The above criterion is based on the assumption that the tracked target from the previous frame and the actual location of the target in the current frame have significant overlap, which is true for most cases.  
	However, such assumption is not always reliable, especially when the camera shifts swiftly. 
	In such cases, we need to match the new observation to the tracked candidates. In Re-ID problems, this is usually accomplished by a visual feature classifier.
	However, visually similar candidates with different identities may confuse such classifiers. Extracting visual features can also be computationally expensive in an online tracking system. Therefore, we design a \textit{Graph Convolution Network }(GCN) to leverage the graphical representation of the human joints. We observe that in two adjacent frames, the location of a person may drift away due to sudden camera shift, but the human pose will stay almost the same as people usually cannot act that fast, as illustrated in Fig. \ref{fig:camera_shift}. Consequently, the graph representation of human skeletons can be a strong cue for candidate matching, which we refer to as pose matching in the following text. 
	
	\subsection{Siamese Graph Convolutional Networks}
	
\begin{figure}
		\vspace{-.25in}
		\centering
		\includegraphics[width=0.95\linewidth]{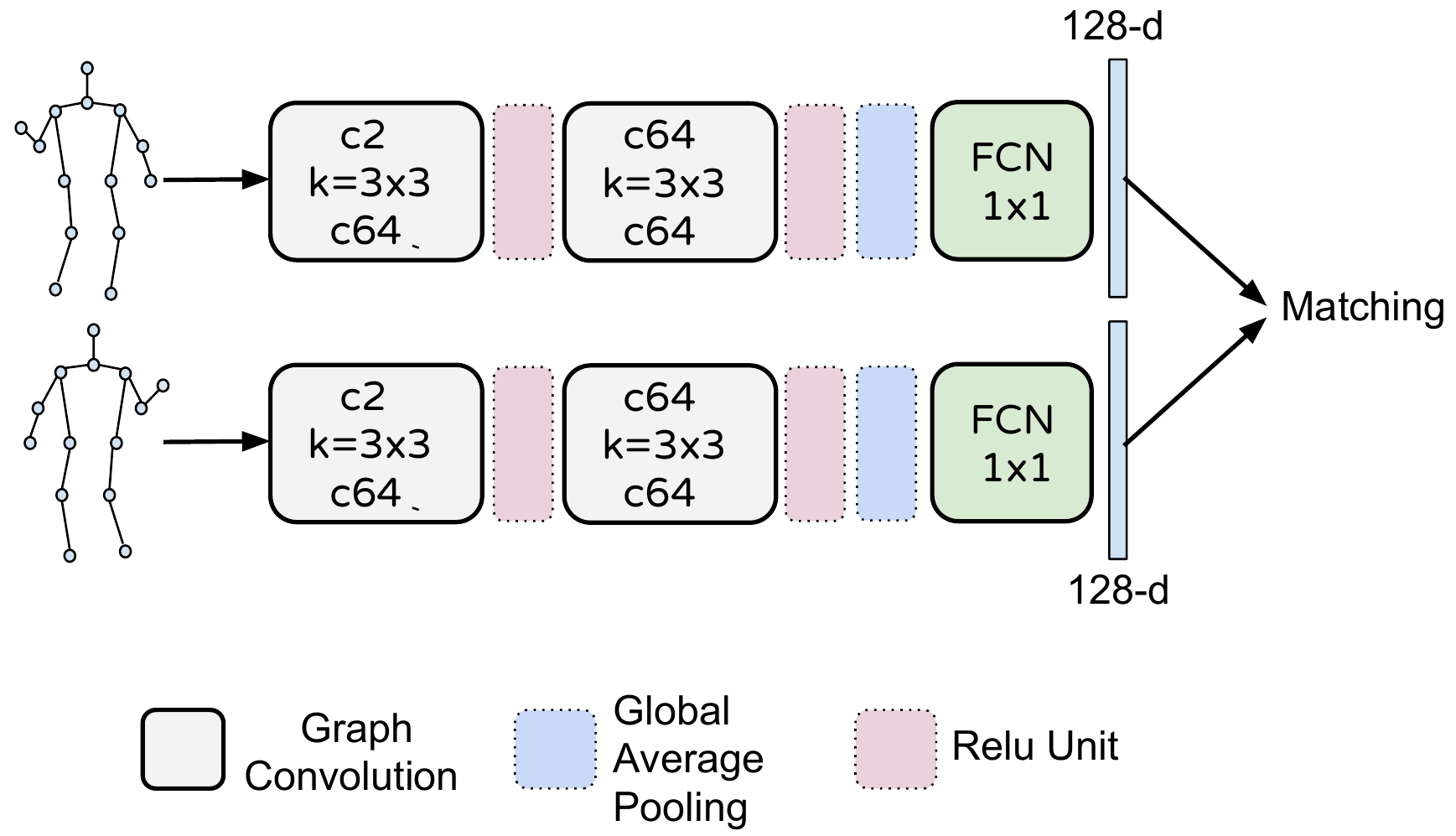}
		\caption{The siamese graph convolution network for pose matching. We extract two feature vectors from the input graph pair with shared network weight. The feature vectors inherently encode the spatial relationship among the human joints.} 
		\label{fig:gcn_network}
		\vspace{-.15in}
	\end{figure}

	\smallskip
	\noindent\textbf{Siamese Network:} 
	Given the sequences of body joints in the form of 2D coordinates, we construct a spatial graph with the joints as graph nodes and connectivities in human body structures as graph edges. 
	The input to our graph convolutional network is the joint coordinate vectors on the graph nodes. It is analogous to image-based CNNs where the input is formed by pixel intensity vectors residing on the 2D image grid \cite{stgcn2018aaai}.
	Multiple graph convolutions are performed on the input to generate a feature representation vector as a conceptual summary of the human pose. It inherently encodes the spatial relationship among the human joints.
	The input to the siamese networks is therefore a pair of inputs to the GCN network. The distance between two output features represent how similar two poses are to each other. Two poses are called a match if they are conceptually similar.  
	The network is illustrated in Fig. \ref{fig:gcn_network}.
	The siamese network consists of $2$ GCN layers and $1$ convolutional layer using contrastive loss. We take normalized keypoint coordinates as input; the output is a $128$ dimensional feature vector. The network is optimized with contrastive loss $\mathcal{L}$ because we want the network to generate feature representations, that are close by enough for positive pairs, whereas they are far away at least by a minimum for negative pairs. we employ the margin contrastive loss: 
	\begin{equation}
	\begin{aligned}
	\mathcal{L}(p_{j},p_{k}, y_{jk})=\frac{1}{2}y_{jk} D^{2}+\frac{1}{2}(1-y_{jk})\max(0,\epsilon-D^{2}),
	\end{aligned}
	\label{eqn:eqn_contrastive_loss}
	\end{equation}
	where $D=\|f(p_{j})-f(p_{k})\|_2$ is the Euclidean distance of two $\ell_2$-norm normalized latent representations, $y_{jk}\in\{0,1\}$ indicates whether $p_{j}$ and $p_{k}$ are the same pose or not, and $\epsilon$ is the minimum distance margin that pairs depicting different poses should satisfy.
	
	\noindent\textbf{Graph Convolution for Skeleton:} 
	For standard 2D convolution on natural images, the output feature maps can have the same size as the input feature maps with stride $1$ and appropriate padding. Similarly, the graph convolution operation is designed to output graphs with the same number of nodes. The dimensionality of attributes of these nodes, which is analogous to the number of feature map channels in standard convolution, may change after the graph convolution operation.
	
	The standard convolution operation is defined as follows:
	given a convolution operator with the kernel size of $ K\times K $, and an input feature map $ f_{in} $ with the number of channels $ c $, the output value of a single channel at the spatial location $ \mathbf{x} $ can be written as:
	\vspace{-3pt}
	\begin{equation}\label{eq:gen_conv}
	f_{out}(\mathbf{x}) = \sum_{h=1}^{K} \sum_{w=1}^{K} f_{in}(\mathbf{s}(\mathbf{x}, h, w))\cdot \mathbf{w}(h, w), 	
	\vspace{-3pt}
	\end{equation}
	where the \textbf{sampling function} $ \mathbf{s}: Z^2\times Z^2\rightarrow Z^2 $ enumerates the neighbors of location $ \mathbf{x} $.
	%In the case of image convolution, it can also be represented as $ \mathbf{s}(\mathbf{x}, h, w) = \mathbf{x} + \mathbf{p^\prime}(h, w). $ 
	The \textbf{weight function} $ \mathbf{w}: Z^2\rightarrow \mathbb{R}^c$ provides a weight vector in $ c $-dimension real space for computing the inner product with the sampled input feature vectors of dimension $ c $.
	%Note that the weight function is irrelevant to the input location $ \mathbf{x} $.
	%Thus the filter weights are shared everywhere on the input image.
	%Standard convolution on the image domain is therefore achieved by encoding a rectangular grid in $ \mathbf{p}(\mathbf{x}) $.
	
\begin{figure}
		\vspace{-.25in}
		\centering
		\includegraphics[width=0.75\linewidth]{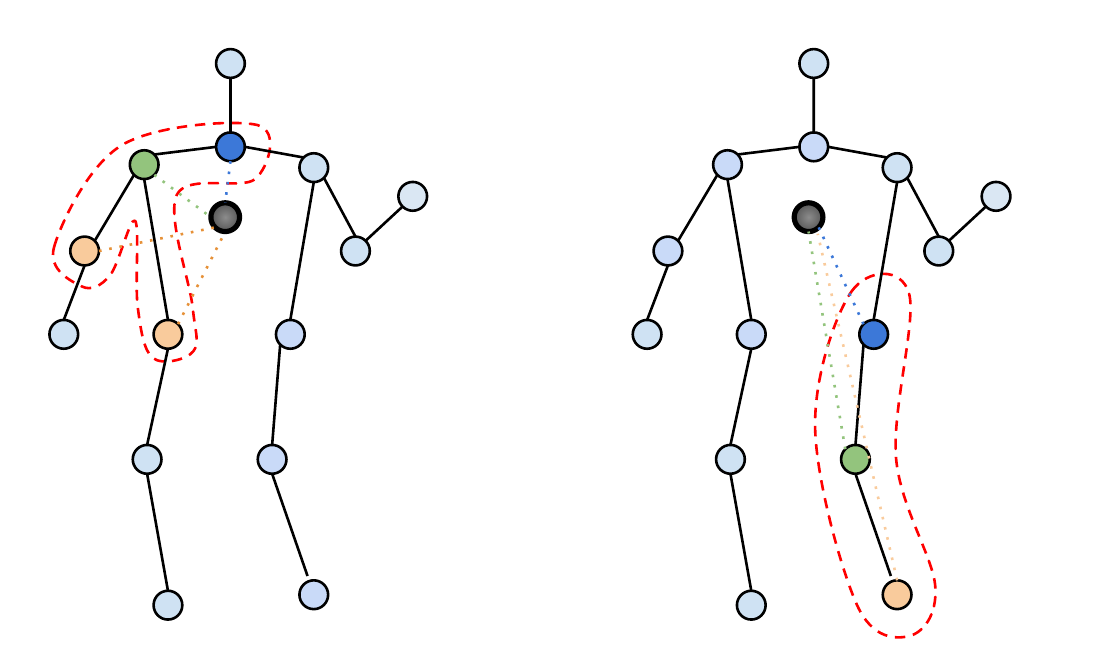}
		\caption{The spatial configuration partitioning strategy proposed in \cite{stgcn2018aaai} for graph sampling and weighting to construct graph convolution operations. The nodes are labeled according to their distances to the skeleton gravity center (black circle) compared with that of the root node (green). Centripetal nodes have shorter distances (blue), while centrifugal nodes have longer distances (yellow) than the root node.} 
		\label{fig:skeleton}
		\vspace{-.15in}
	\end{figure}

	The convolution operation on graphs is defined by extending the above formulation to the cases where the input features map resides on a spatial graph $ V_t$, \emph{i.e.} the feature map $ f_{in}^t : V_t\rightarrow R^c$ has a vector on each node of the graph.
	The next step of the extension is to re-define the sampling function $ \mathbf{p} $ and the weight function $ \mathbf{w} $. We follow the method proposed in  \cite{stgcn2018aaai}.
	For each node, only its adjacent nodes are sampled. The neighbor set for node $v_{i}$ is $B(v_{i}) = \{ v_{j} | d(v_{j}, v_{i}) \leq  1 \}$.
	The sampling function $ \mathbf{p}:B(v_{i})\rightarrow V $ can be written as $\mathbf{p}(v_{i}, v_{j}) = v_{j}$.
	In this way, the number of adjacent nodes is not fixed, nor is the weighting order.
	In order to have a fixed number of samples and a fixed order of weighting them, we label the neighbor nodes around the root node with fixed number of partitions, and then weight these nodes based on their partition class. 
	The specific partitioning method is illustrated in Fig. \ref{fig:skeleton}.
	%More detailed explanation can be found in~\cite{stgcn2018aaai}.
	
	Therefore, Eq.~(\ref{eq:gen_conv})  for graph convolution is re-written as:
	\vspace{-3pt}\begin{align}
	\label{eq:graph_conv}
	f_{out}(v_{i}) = \sum_{v_{j}\in B(v_{i})} \frac{1}{Z_{i}(v_{j})}f_{in}(\mathbf{p}(v_{i}, v_{j}))\cdot \mathbf{w}(v_{i}, v_{j}), 
	\vspace{-3pt}\end{align}
	where the normalization term 
	$
	Z_{i}(v_{j}) = \mid\{ v_{k} |  l_{i}(v_{k}) = l_{i}(v_{j})\}\mid
	$
	is to balance the contributions of different subsets to the output.
	According to the partition method mentioned above, we have:
	\vspace{-3pt}\begin{equation}
	l_{i}({v_j})=
	\begin{cases}
	0 &\mbox{if $r_{j} = r_{i}$}\\
	1 &\mbox{if $r_{j} < r_{i}$}\\
	2 &\mbox{if $r_{j} > r_{i}$}
	\end{cases}
	\vspace{-3pt}\end{equation}
	where $r_i$ is the average distance from gravity center to joint $i$ over all frames in the training set.
	
	%-------------------------------------------------------------------------
	\section{Experiments}
\begin{figure*}
		\vspace{-.2in}
		\centering
		\includegraphics[width=0.85\linewidth]{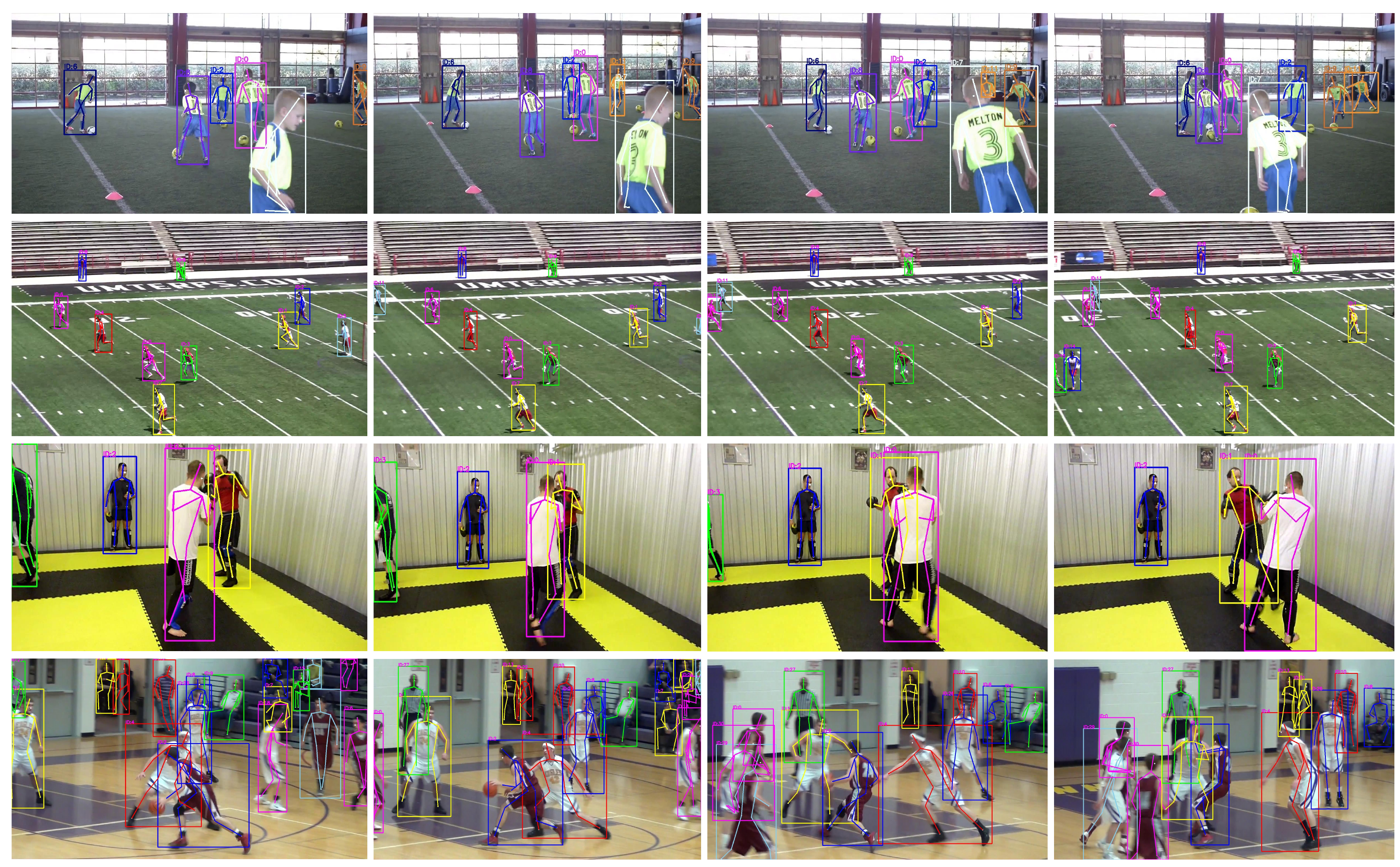}
		\caption{Qualitative evaluation results. Each person is visualized with a different color. Same color indicates identical IDs.} 
		\label{fig:quality}
		\vspace{-.1in}
	\end{figure*}

	In this section, we present quantitative results of our experiments. Some qualitative results are shown in Fig. \ref{fig:quality}.
	
	\subsection{Dataset}
	
	PoseTrack \cite{andriluka2018posetrack} is a large-scale benchmark for human pose estimation and articulated tracking in videos. It provides publicly available training and validation sets as well as an evaluation server for benchmarking on a held-out test set. The benchmark is a basis for the challenge competitions at ICCV'17 \cite{PoseTrack17} and ECCV'18 \cite{PoseTrack18} workshops.
	%The ICCV'17 workshop is organized around a challenge with three competition tracks focusing on single frame multi-person pose estimation, multi-person pose estimation in videos, and multi-person articulated tracking.
	The dataset consisted of over $68,000$ frames for the ICCV'17 challenge and is extended to twice as many frames for the ECCV'18 challenge. It now includes $593$ training videos, $74$ validation videos and $375$ testing videos. 
	%The ECCV'18 workshop is organized around a challenge with two competition tracks focusing on multi-person pose estimation in videos, and multi-person articulated tracking. 
	%In this work, we focus on multi-person pose tracking, perform experiments on the validation sets of both PoseTrack'17 and PoseTrack'18.
	%This dataset can be used to evaluate approaches on either validation set or held-out test set. 
	For held-out test set, at most four submissions per task can be made for the same approach. 
	%Any two submissions should be 72 hours apart. It is advised to evaluate approaches on validation data first to avoid losing evaluation attempts due to formatting mismatch or other problems. 
	Evaluation on validation set has no submission limit. Therefore, ablation studies in Section \ref{sec:ablation} are performed on the validation set.
	Since PoseTrack'18 test set is not open yet, we compare our results with other approaches in Sec. \ref{sec:comparison} on PoseTrack'17 test set. 
	
	\subsection{Evaluation  Metrics}
	The evaluation includes pose estimation accuracy and pose tracking
	accuracy. Pose estimation accuracy is evaluated using the standard \textbf{mAP} metric, whereas the evaluation of pose tracking is according to the \textbf{clear MOT} \cite{bernardin2008evaluating} metrics that are the standard for evaluation of multi-target tracking. 
	%In addition, trajectory-based measures are also evaluated that count the number of mostly tracked (MT), mostly lost (ML) tracks and the number of times a ground-truth trajectory is fragmented (FM). 
	
	\subsection{Implementation  Details}
	\label{sec:details}
	We adopt state-of-the-art key-frame object detectors trained with ImageNet and COCO datasets. Specifically, we use pre-trained models from deformable ConvNets \cite{dai17dcn}. 
	We conduct experiments on validation sets to choose the object detector with better recall rates.
	For the object detectors, we compare the deformable convolution versions of the R-FCN network \cite{dai2016r} and of the FPN network \cite{lin2017feature}, both with ResNet101 backbone \cite{he2016deep}.
	The FPN feature extractor is attached to the Fast R-CNN \cite{girshick2015fast} head for detection.
	We compare the detection results with the ground truth based on the precision and recall rate on PoseTrack'17 validation set.  
	In order to eliminate redundant candidates, we drop candidates with lower {likelihood}. 
	As shown in Table \ref{table_detectors}, precision and recall of the detectors are given for various drop thresholds. Since the FPN network performs better, we choose it as our human candidate detector. 
	During training, we infer ground truth bounding boxes of candidates from the annotated keypoints, because in PoseTrack'17 dataset, the bounding box positions are not provided in the annotations. Specifically, we locate a bounding box from the minimum and maximum coordinates of the $15$ keypoints, and then enlarge this box by 20\% both horizontally and vertically. %Even though ground truth bounding boxes are given in PoseTrack 2018 dataset, we infer a more consistent version based on ground truth locations of keypoints. Those inferred ground truth bounding boxes are utilized to train the pose estimator.
	
	For the single-person human pose estimator, we adopt CPN101 \cite{chen2017cascaded} and MSRA152 \cite{xiao2018simple} with slight modifications. 
	We first train the networks with the merged dataset of PoseTrack'17 and COCO for 260 epochs. Then we finetune the network solely on PoseTrack'17 for 40 epochs in order to mitigate the inaccurate regression on head and neck. For COCO, bottom-head and top-head positions are not given. We infer these keypoints by interpolation on the annotated keypoints. We find that by finetuning on the PoseTrack dataset, the prediction on head keypoints will be refined. During finetuning, we use the technique of online hard keypoint mining, only focusing on losses from the $7$ hardest keypoints out of the total $15$ keypoints. 
	Pose inference is performed online with single thread.
	
	For the pose matching module, we train a siamese graph convolutional network with $2$ GCN layers and $1$ convolutional layer using contrastive loss. 
	%(We also tried triplet loss, which does not render better performance.) 
	We take normalized keypoint coordinates as input; the output is a $128$ dimensional feature vector. 
	Following \cite{stgcn2018aaai}, we use spatial configuration partitioning as the sampling method for graph convolution and use learnable edge importance weighting. 
	To train the siamese network, we generate training data from the PoseTrack dataset. Specifically, we extract people with same IDs within adjacent frames as positive pairs, and extract people with different IDs within the same frame and across frames as negative pairs. Hard negative pairs only include spatially overlapped poses. 
	The number of collected pairs are illustrated in Table \ref{pose_pairs}.
	We train the model with batch size of $32$ for a total of $200$ epochs with SGD optimizer. Initial learning rate is set to $0.001$ and is decayed by $0.1$ at epochs of $40, 60, 80, 100$. Weight decay is $10^{-4}$.
	\begin{table}[!t]
		\vspace{-1pt}
		\setlength{\tabcolsep}{8pt}
		\small
		\centering 
		\begin{tabular}{l c c}
			\toprule
			-  & Train  & Validation \\
			\midrule
			Positive Pairs         & 56908  & 9731 \\
			Hard Negative Pairs  & 25064  & 7020  \\
			Other Negative Pairs       & 241450  & 91228 \\
			\bottomrule
		\end{tabular}
		\vspace{3pt}
		\caption{Pose pairs collected from PoseTrack'18 dataset. }
		\label{pose_pairs}
		\vspace{-.2in}
	\end{table}

	\subsection{Ablation Study}
	\label{sec:ablation}
	We conducted a series of ablation studies to analyze the contribution of each component on the overall performance. 
	
	\begin{table}[h]
		\vspace{-3pt}
		\setlength{\tabcolsep}{3pt}
		%\small
		\footnotesize
		\centering 
		%\begin{tabular}{l|l|l|l|l|l|l}
		\begin{tabular}{llccccc}
			\toprule
			- &Method / Thresh  &  0.1 & 0.2 & 0.3 & \textbf{0.4} & 0.5 \\%& 0.6 & 0.7 & 0.8 & 0.9 \\
			\midrule			\multirow{2}{*}{\textbf{Prec}} 
			&Deformable FPN & 17.9 & 27.5 &32.2 &\textbf{34.2}  &35.7 \\%&37.2 &38.6 &40.0 &42.1 \\
			&Deformable R-FCN & 15.4 & 21.1 &25.9 &30.3  &34.5 \\%&37.9 &39.9 &41.6 &43.2 \\
			\midrule
			\multirow{2}{*}{\textbf{Recall}}
			&Deform FPN & 87.7 &86.0 &84.5  &\textbf{83.0} &80.8 \\%&79.2 &77.0 &73.8 &69.0 \\
			&Deform R-FCN & 87.7 &86.5 &85.0  &82.6 &80.1 \\%&77.3 &74.4 &70.4 &61.0 \\
			\bottomrule
		\end{tabular}
		\vspace{3pt}
		\caption{Comparison of detectors: Precision-Recall on PoseTrack 2017 validation set. A bounding box is correct if its IoU with GT is above certain threshold, which is set to 0.4 for all experiments.}
		\label{table_detectors}
		\vspace{-.2in}
	\end{table}

	\begin{table}[h]
		\setlength{\tabcolsep}{4pt}
		\footnotesize
		\centering 
		\begin{tabular}{lcccccc}
			\toprule
			\multicolumn{1}{l}{ - }  
			&
			\multicolumn{3}{c}{ \textbf{Estimation (mAP) }}  
			&                                            
			\multicolumn{3}{c}{ \textbf{Tracking (MOTA)}} \\\midrule
			Method & Wri  & Ankl & Total  & Wri  & Ankl & Total \\
			
			\midrule
			GT Detections & 74.7  & 75.4 & \textbf{81.7} & 56.3 & 56.2 & \textbf{67.0} \\
			Deform FPN-101  & 70.2  & 64.7 &\textbf{ 74.6 } & 54.6 & 48.7 & \textbf{61.3} \\
			Deform RFCN-101 & 69.0 & 64.3 & \textbf{73.7}  & 52.2 & 47.4 & \textbf{59.0} \\
			\bottomrule
		\end{tabular}
		\vspace{3pt}
		\caption{Comparison of offline pose tracking results using various detectors on PoseTrack'17 validation set. }
		\label{table-gap-task1and3}
		\vspace{-.1in}
	\end{table}
	
	\noindent\textbf{Detectors:} 
	We experimented with several detectors and decide to use Deformable ConvNets with ResNet101 as backbone, \textit{Feature Pyramid Networks} (FPN) for feature extraction, and fast R-CNN scheme as detection head. As shown in Table \ref{table_detectors}, this detector performs better than Deformable R-FCN with the same backbone. It is no surprise that the better detector results in better performances on both pose estimation and pose tracking, as shown in Table \ref{table-gap-task1and3}.

	\noindent\textbf{Offline vs. Online:} 
	We studied the effect of keyframe intervals of our online method and compare with the offline method. 
	%We report the ablation experiments on tracking performance evaluated using Multiple Object Tracking Accuracy (MOTA) metric. 
	For fair comparison, we use identical human candidate detector and pose estimator for both methods. For offline method, we pre-compute human candidate detection and estimate the pose for each candidate, then we adopt a flow-based pose tracker \cite{xiu2018pose}, where pose flows are built by associating poses that indicate the same person across frames.
	%We start the tracking process from the first frame where human candidates are detected. 
	For online method, we perform truly online pose tracking. Since human candidate detection is performed only at key frames, the online performance varies with different intervals.
	In Table \ref{offline-online}, we illustrate the performance of the offline method, compared with the online method that is given various keyframe intervals. Offline methods performed better than online methods. But we can see the great potential of online methods when the detections (DET) at keyframes are more accurate, the upper-limited of which is achieved with ground truth (GT) detections. As expected, frequent keyframe helps more with the performance. 
	Note that the online methods only use spatial consistency for data association at key frames. We report ablation experiments on the pose matching module in the following text.
	
	\begin{table}[h]
		\vspace{-3pt}
		\setlength{\tabcolsep}{5pt}
		\footnotesize
		\centering 
		\begin{tabular}{lcccccc}
			\toprule
			\multicolumn{1}{l}{ - }  
			&
			\multicolumn{3}{c}{ \textbf{Estimation (mAP) }}  
			&                                            
			\multicolumn{3}{c}{ \textbf{Tracking (MOTA)}} \\\midrule
			\textbf{Method} & Wri  & Ankl & Total  & Wri  & Ankl & Total \\
			\midrule
			Offline-CPN101  &72.6 &68.9 &\textbf{76.4}  &56.1 &55.3 &\textbf{62.4}\\
			Offline-MSRA152  &73.6 &70.5 &\textbf{77.3}  &58.5 &58.5 &\textbf{64.9}\\
			\midrule
			Online-DET-CPN101-8F  &70.5 &68.3 &\textbf{74.0}  &52.4 &50.3 &\textbf{58.1}\\
			Online-DET-CPN101-5F   &71.7  &68.9  &\textbf{75.1}  &53.3 &51.0 &\textbf{59.0}\\
			Online-DET-CPN101-2F   &72.4 &69.1 &\textbf{76.0}  &54.2 &51.5 &\textbf{60.0}\\
			\midrule
			Online-DET-MSRA152-8F  &71.1 &69.5 &\textbf{75.0}  &54.6 &54.6 &\textbf{61.0}\\
			Online-DET-MSRA152-5F   &72.1 & 70.4 &\textbf{76.1}  &55.2 &55.5 &\textbf{61.9}\\
			Online-DET-MSRA152-2F   &73.3 & 70.9 &\textbf{77.2}  &56.5 &56.6 &\textbf{63.3}\\
			\bottomrule
		\end{tabular}
		\vspace{3pt}
		\caption{Comparison of offline and online pose tracking results with various keyframe intervals on PoseTrack'18 validation set. }
		\label{offline-online}
		\vspace{-.2in}
	\end{table}

\begin{figure}[h]
		\centering
		\includegraphics[width=0.95\linewidth]{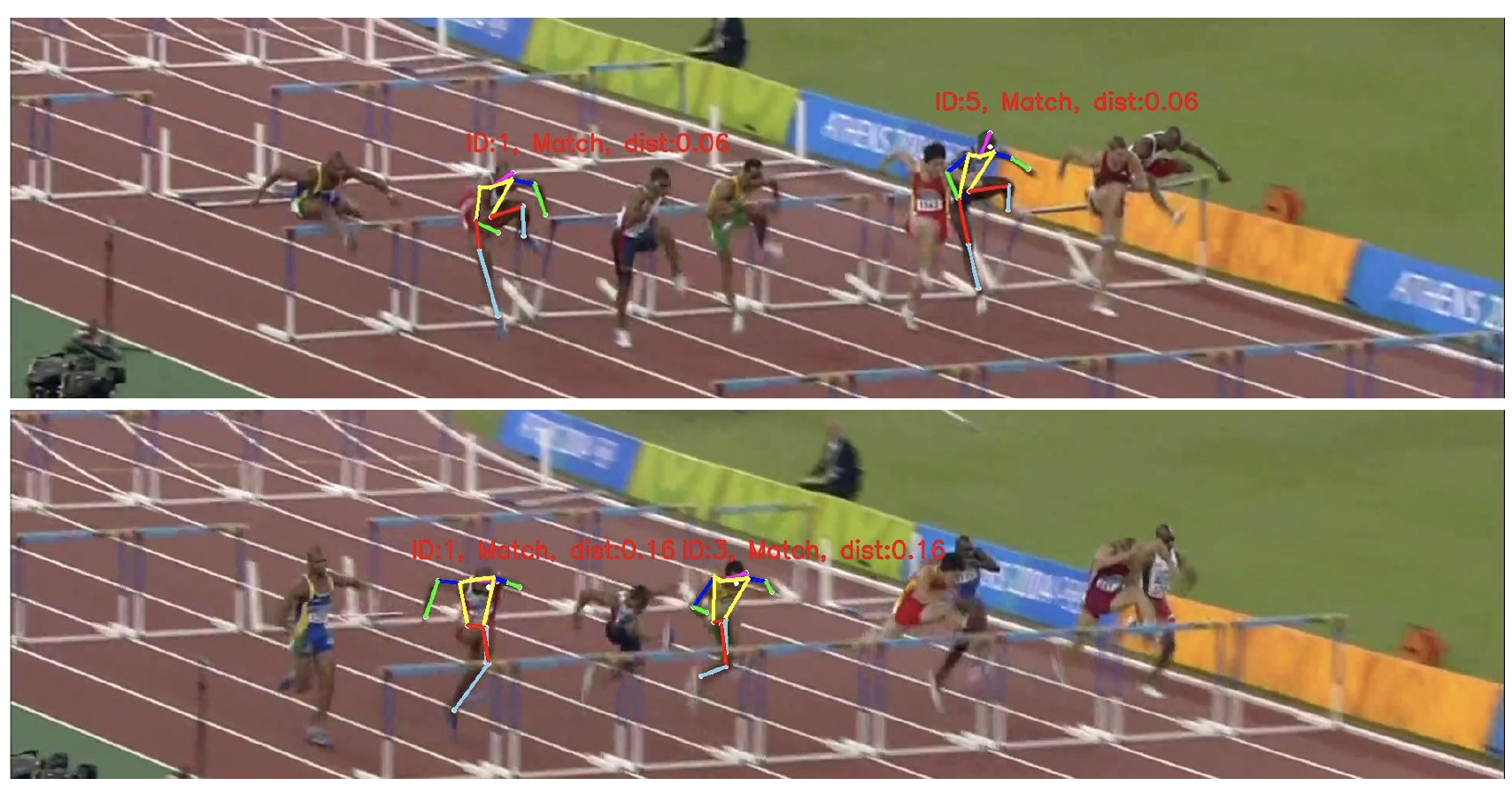}
		\caption{In some situations, different people indeed have very similar poses. Therefore, spatial consistency is considered first.} 
		\label{fig:similarity}
		\vspace{-.1in}
	\end{figure}

	\noindent\textbf{GCN vs. Spatial Consistency (SC):} Next, we report results when pose matching is performed during data association stage, compared with only employing spatial consistency. 
	It can be shown in Table \ref{GCN_SC} that the tracking performance increases with GCN-based pose matching. 
	However, in some situations, different people may have near-duplicate poses, as shown in Fig. \ref{fig:similarity}. To mitigate such ambiguities, spatial consistency is considered prior to pose similarity.

	\begin{table}[h]
		%\small
		\vspace{-3pt}
		\setlength{\tabcolsep}{5pt}
		\footnotesize
		\centering 
		\begin{tabular}{lccccc}
			\toprule
			\multirow{2}{*}{\textbf{Method}} &\multirow{2}{*}{\textbf{Detect}} &\multirow{2}{*}{\textbf{Keyframe}}  &\multicolumn{2}{c}{\textbf{MOTA}}  \\
			\cline{4-5}
			& & &CPN101 &MSRA152 \\
			\midrule
			SC        &\multirow{6}{*}{GT} &\multirow{2}{*}{8F}  &68.2 &72.0   \\
            SC+GCN & & &68.9 & 72.6    \\
            \cmidrule{1-1}\cmidrule{3-5}
            SC       & &\multirow{2}{*}{5F}  &68.7 &73.0  \\
            SC+GCN & &  & 69.2  &73.5   \\
             \cmidrule{1-1}\cmidrule{3-5}
            SC       & &\multirow{2}{*}{2F}  &72.0 &76.7   \\
            SC+GCN & &  & 73.5  &78.0   \\
			\midrule
			SC        &\multirow{6}{*}{DET} &\multirow{2}{*}{8F}  &58.1 &61.0   \\
			SC+GCN & & &59.0 & 62.1    \\
			\cmidrule{1-1}\cmidrule{3-5}
			SC       & &\multirow{2}{*}{5F}  &59.0 &61.9  \\
			SC+GCN & &  & 60.1  &63.1   \\
			\cmidrule{1-1}\cmidrule{3-5}
		    SC       & &\multirow{2}{*}{2F}  &60.0 &63.3   \\
	        SC+GCN & &  & 61.3  &64.6   \\
			\bottomrule
		\end{tabular}
		\vspace{3pt}
		\caption{Performance comparison of LightTrack with GCN and SC on PoseTrack'18 validation set.}
		\vspace{-.2in}
		\label{GCN_SC}
	\end{table}
	%\smallskip

	%\smallskip
	\noindent\textbf{GCN vs. Euclidean Distance (ED):} We studied whether the GCN network outperforms naive pose matching scheme. With same normalization on the keypoints, ED as the dissimilarity metric for pose matching renders 85\% accuracy on validation pairs generated from PoseTrack dataset, while GCN renders 92\% accuracy. 
	We validate on positive pairs and hard negative pairs.
	%The positive pairs in the validation set contain pose pairs where the poses have same ID with or without bounding box overlap in adjacent frames; the negative data consists of pairs whose poses have overlapping bounding boxes but different IDs.

	\subsection{Performance Comparison}
	\label{sec:comparison}
	Since PoseTrack'18 test set is not open yet, we compare our methods with other approaches, both online and offline, on PoseTrack'17 test set. 
	For fair comparison, we only use PoseTrack'17 training set and COCO train+val set to train the pose estimators. No auxiliary data is used. 
	We performed ablation studies on validation sets with CPN-101 \cite{chen2017cascaded} as the pose estimator. During testing, in addition to CPN-101, we conduct experiments using MSRA-152 \cite{xiao2018simple}.
	
	\begin{table}[h]
		\vspace{-3pt}
		\begin{center}
			\hspace{-0.33in}			\resizebox{0.52\textwidth}{!}{
				\setlength{\tabcolsep}{4pt}
				\begin{tabular}{c l | c c |c |c |c}
					\cmidrule{2-7}
					& \textbf{Method} & \textbf{Wrist-AP} & \textbf{Ankles-AP} & \textbf{mAP} & \textbf{MOTA} & \textbf{fps} \\
					\cmidrule{2-7}
					\multirow{7}{*}{\rotatebox[origin=c]{90}{~\textbf{Offline}}} &\multicolumn{5}{ c }
					{Posetrack 2017 Test Set} \\
					\cmidrule{2-7}
					&PoseTrack, CVPR'18 \cite{andriluka2018posetrack} &54.3 &49.2 &59.4 &48.4 &-\\
					&BUTD, ICCV'17 \cite{jin2017towards} &52.9 &42.6 &59.1 &50.6 &-\\
					&Detect-and-track, CVPR'18 \cite{girdhar2018detect} &- &- &59.6 &51.8 & -\\
					&Flowtrack-152, ECCV'18 \cite{xiao2018simple} &71.5 &65.7 &74.6 &57.8 &-\\
					&HRNet, CVPR'19\cite{sun2019deep}  &72.0 &67.0 &74.9 &57.9 &-\\
					
					&\textbf{Ours-CPN101 (offline)} & \textbf{68.0 / 59.7} & \textbf{62.6 / 56.3} & \textbf{70.7 / 63.9}& \textbf{55.1 }& -\\
					&\textbf{Ours-MSRA152 (offline)} & \textbf{68.9 / 61.8} & \textbf{63.2 / 58.4} & \textbf{71.5 / 65.7} & \textbf{57.0} & -\\
					&\textbf{Ours-manifold (offline)} & \textbf{- / 64.6} & \textbf{- / 58.4} & \textbf{- / 66.7} & \textbf{58.0} & -\\
					\midrule
					\multirow{4}{*}{\rotatebox[origin=c]{90}{~\textbf{Online}}}
					
					&PoseFlow, BMVC'18 \cite{xiu2018pose} &59.0 &57.9 &63.0 &51.0 &10*\\
					&JointFlow, Arxiv'18 \cite{doering2018joint} &53.1 &50.4 &63.3 &53.1 &0.2\\
					&\textbf{Ours-CPN101-LightTrack-3F} &\textbf{61.2} &\textbf{57.6} &\textbf{63.8 }&\textbf{52.3} &\textbf{47* / 0.8}\\
					&\textbf{Ours-MSRA152-LightTrack-3F} & \textbf{63.8} & \textbf{59.1} & \textbf{66.5} & \textbf{55.1} & \textbf{48* / 0.7}\\
					
					\cmidrule{2-7}
					\multirow{5}{*}{\rotatebox[origin=c]{90}{~}}&\multicolumn{5}{ c }
					{Posetrack 2018 Validation Set} \\
					\cmidrule{2-7}
					&Ours-CPN101 (offline) & 72.6 / 63.9 & 68.9 / 62.6 & 76.4 / 69.7 & 62.4 & -\\
					&Ours-MSRA152 (offline) & 73.6 / 65.6 & 70.5 / 64.9 & 77.3 / 71.2 & 64.9 & -\\
					\cmidrule{2-7}
					 &Ours-YoloMD-LightTrack-2F & 62.9 / 56.2 & 57.8 / 53.3 & 70.4 / 66.0 & 55.7 & 59* / 1.9 \\
					 &Ours-CPN101-LightTrack-2F & 72.4 / 66.3  & 69.1 / 64.2 & 76.0 / 70.3 & 61.3 &47* / 0.8\\
					 &Ours-MSRA152-LightTrack-2F & 73.3 / 66.4 & 70.9 / 66.1 & 77.2 / 72.4 & 64.6 & 48* / 0.7 \\
					\cmidrule{2-7}
			\end{tabular}}
		\end{center}
		\vspace{-.1in}
		\caption{
			Performance comparison on Posetrack dataset. The last column shows the speed in frames per second (* means excluding pose inference time). For our online methods, mAP are provided after keypoints dropping. For our offline methods, mAP are provided both before (left) and after (right) keypoints dropping.
		}
		\vspace{-.1in}
		\label{table:PoseTrackComparison}
	\end{table}

	%\smallskip
	\noindent\textbf{Accuracy:} 
	As shown in Table \ref{table:PoseTrackComparison}, our method LightTrack outperforms other online methods while maintaining a much higher frame rate, and is very competitive with offline state-of-the-arts. 
	For our offline approach, we use the same detector and pose estimator of LightTrack, except we replace LightTrack with the official release of PoseFlow \cite{xiu2018pose} for performance comparison. Although the PoseFlow algorithm is conceptually online, the processing is performed in multiple stages, and requires keypoint-matching between frames pre-computed, which is computationally expensive. In contrast, our LightTrack is truly processed online.

	\noindent\textbf{Speed:} 
	Testing on single Telsa P40 GPU, pose matching costs an average of $2.9$ ms for each pair.
	Since pose matching only occurs at key-frames, its frequency of occurrence depends on the number of candidates and length of keyframe intervals. 
	Therefore, we test the average processing time on the PoseTrack'18 validation set, which consists of $74$ videos with a total of $8,857$ frames. It takes the online algorithm CPN101-LightTrack $11,638$ seconds to process, of which $11,450$ secs used for pose estimation. The frame rate of the whole system is $0.76$ fps.  
	The framework runs at around $47.11$ fps excluding pose inference time.
	In total, $57,928$ persons are encountered. An average of $6.54$ people are tracked for each frame.
	It takes CPN101 $140$ ms to process each human candidate, including $109$ ms pose inference and $31$ ms for pre-processing and post-processing.  
	There is potential room for the actual frame rate and tracking performance to improve with other choices of pose estimators and parallel inference optimization.
	We see an improved performance with MSRA152-LightTrack but slightly slower frame rate due to its $133$ ms inference time. 
	
	\subsection{Discussions}
	%\smallskip
	\noindent\textbf{Accuracy:} 
	Since the components in our framework are easily replaceable and extendable, methods employing this framework can potentially become faster, more accurate, or possibly both.
	Note that the pose estimator mentioned in section \ref{sec:details} can be replaced by a more accurate \cite{li2019rethinking} or a much faster counterpart. 
	The performance boost in the general object detector, or methods that focus on detecting people (\emph{e.g.}, using auxiliary dataset \cite{li2018crowdpose}), should also improve the pose tracking performance.
	Ablation study in section \ref{sec:ablation} has shown that better detection increases the MOTA score, regardless of which detectors to use.

	\noindent\textbf{Speed:} 
	The pose estimation network can be prioritized for speed while sacrificing some accuracy. 
	For instance, we use YOLOv3 and MobileNetv1-deconv (YoloMD) as detector and pose estimator, respectively. It achieves an average of $2$ FPS with $70.4$ mAP and MOTA score $55.7$\% on PoseTrack'18 validation set.  
	Aside from network structure design, a faster network could also refine heatmaps from previous frame(s). 
	Recently, refinement-based networks \cite{moon2018posefix, Fieraru2018LearningTR} have drawn enormous attention. 
	%Intuitively, the sub-network after intermediate supervision can be separated as a new pose estimation network that refines upon any existing pose estimator. 
	%It is both desirable and feasible to place extra emphasis on the speed with pre-computed predictions considered as knowledge prior.

	\noindent\textbf{Flexibility:} 
	The advantage of our top-down approach in pose tracking is that we can conveniently track specific targets and do not necessarily track all candidates. It can be achieved simply by choosing the target(s) at the first frame and providing target locations at key-frames. As a side effect, this further reduces computational complexity. If the target has specific visual appearance, the framework can be conveniently extended to ensure only the target can be matched at key-frames and tracked at remaining frames.

	%------------------------------------------------------------------------
	\section{Conclusions}
	In this paper, we propose an effective and generic light-weight framework for online human pose tracking.
	We also provide a baseline employing this framework, and propose a siamese graph convolution network for human pose matching as a Re-ID module in our pose tracking system. 
	The skeleton-based representation effectively captures human pose similarity and is computationally inexpensive. 
	Our method outperforms other online methods significantly, and is very competitive with offline state-of-the-arts but with much higher frame rate.
	We believe the proposed framework is worthy to be widely used due to its superior performance, generality, and extensibility. 
	%In our future research, we will study the design of the pose estimation module with temporal refinement for more efficient pose tracking. 
	%We will also explore skeleton-based pose matching in spatio-temporal domain.

	%\newpage
	{\small
		\bibliographystyle{ieee}
		\bibliography{ning}
	}

\end{document}